\documentclass[letterpaper]{article} 
\usepackage{aaai2026}  
\usepackage{times}  
\usepackage{helvet}  
\usepackage{courier}  
\usepackage[hyphens]{url}  
\usepackage{graphicx} 
\usepackage{natbib}  
\usepackage{caption} 
\usepackage{algorithm}
\usepackage{algorithmic}
\usepackage{multirow}
\usepackage{booktabs}
\usepackage{amssymb}
\def \ie{\emph{i.e.}}

\def \etal{{\em \etal~}}

\usepackage{newfloat}
\usepackage{listings}
\usepackage{amsmath}
\DeclareCaptionStyle{ruled}{labelfont=normalfont,labelsep=colon,strut=off} 
\floatstyle{ruled}
\newfloat{listing}{tb}{lst}{}
\floatname{listing}{Listing}
\title{Burst Image Quality Assessment: A New Benchmark and Unified Framework for Multiple Downstream Tasks}
\author{
    Xiaoye Liang\textsuperscript{\rm 1},
    Lai Jiang\textsuperscript{\rm 1},
    Minglang Qiao\textsuperscript{\rm 1},
    Yichen Guo\textsuperscript{\rm 1},
    Yue Zhang\textsuperscript{\rm 1}\thanks{Corresponding author},\\
    Xin Deng\textsuperscript{\rm 1},
    Shengxi Li\textsuperscript{\rm 1},
    Yufan Liu\textsuperscript{\rm 2},
    Mai Xu\textsuperscript{\rm 1}
}
\affiliations{

    \textsuperscript{\rm 1}Beijing University of Aeronautics and Astronautics\\
    \textsuperscript{\rm 2}Institute of Automation, Chinese Academy of Sciences

    \{xiaoyeliang,yue\_zhang\}@buaa.edu.cn
}

\begin{document}

\maketitle

\begin{abstract}
In recent years, the development of burst imaging technology has improved the capture and processing capabilities of visual data, enabling a wide range of applications. However, the redundancy in burst images leads to the increased storage and transmission demands, as well as reduced efficiency of downstream tasks. To address this, we propose a new task of Burst Image Quality Assessment (BuIQA), to evaluate the task-driven quality of each frame within a burst sequence, providing reasonable cues for burst image selection. Specifically, we establish the first benchmark dataset for BuIQA, consisting of $7,346$ burst sequences with $45,827$ images and $191,572$ annotated quality scores for multiple downstream scenarios.
Inspired by the data analysis, a unified BuIQA framework is proposed to achieve an efficient adaption for BuIQA under diverse downstream scenarios. 
Specifically, a task-driven prompt generation network is developed with heterogeneous knowledge distillation, to learn the priors of the downstream task. Then, the task-aware quality assessment network is introduced to assess the burst image quality based on the task prompt. Extensive experiments across 10 downstream scenarios demonstrate the impressive BuIQA performance of the proposed approach, outperforming the state-of-the-art. Furthermore, it can achieve $0.33$ dB PSNR improvement in the downstream tasks of denoising and super-resolution, by applying our approach to select the high-quality burst frames.
\end{abstract}


\section{Introduction}
In recent years, burst imaging, a technique that rapidly captures multiple high-resolution frames in quick succession, has revolutionized visual data acquisition and processing, enabling unprecedented precision in both subjective and objective scene analysis. For subjective use, it helps capture the crucial shot of fleeting moments, such as sports events and wildlife photography. More importantly, for objective use, burst imaging enhances computational photography tasks, such as denoising and super-resolution,  by merging multiple frames into a single high-quality output. However, unlike single images, burst images exhibit significant redundancy, leading to 1) increased storage and bandwidth costs, and 2) reduced efficiency in downstream processing tasks. Therefore, there exists a critical need for automatic key frame selection. For example, in album highlight generation, the most visually appealing and memorable frames are saved to reduce storage. Similarly, using a subset of the frames with minimal blur and noise can further improve the output quality of the downstream objective tasks like super-resolution. Therefore, in this paper, we propose a novel task of Burst Image Quality Assessment (BuIQA), which evaluates the subjective or objective quality of the individual frame within a burst sequence, enabling adaptive frame selection for specific downstream tasks.

\begin{figure}[]
\centering
\includegraphics[width=1\columnwidth]{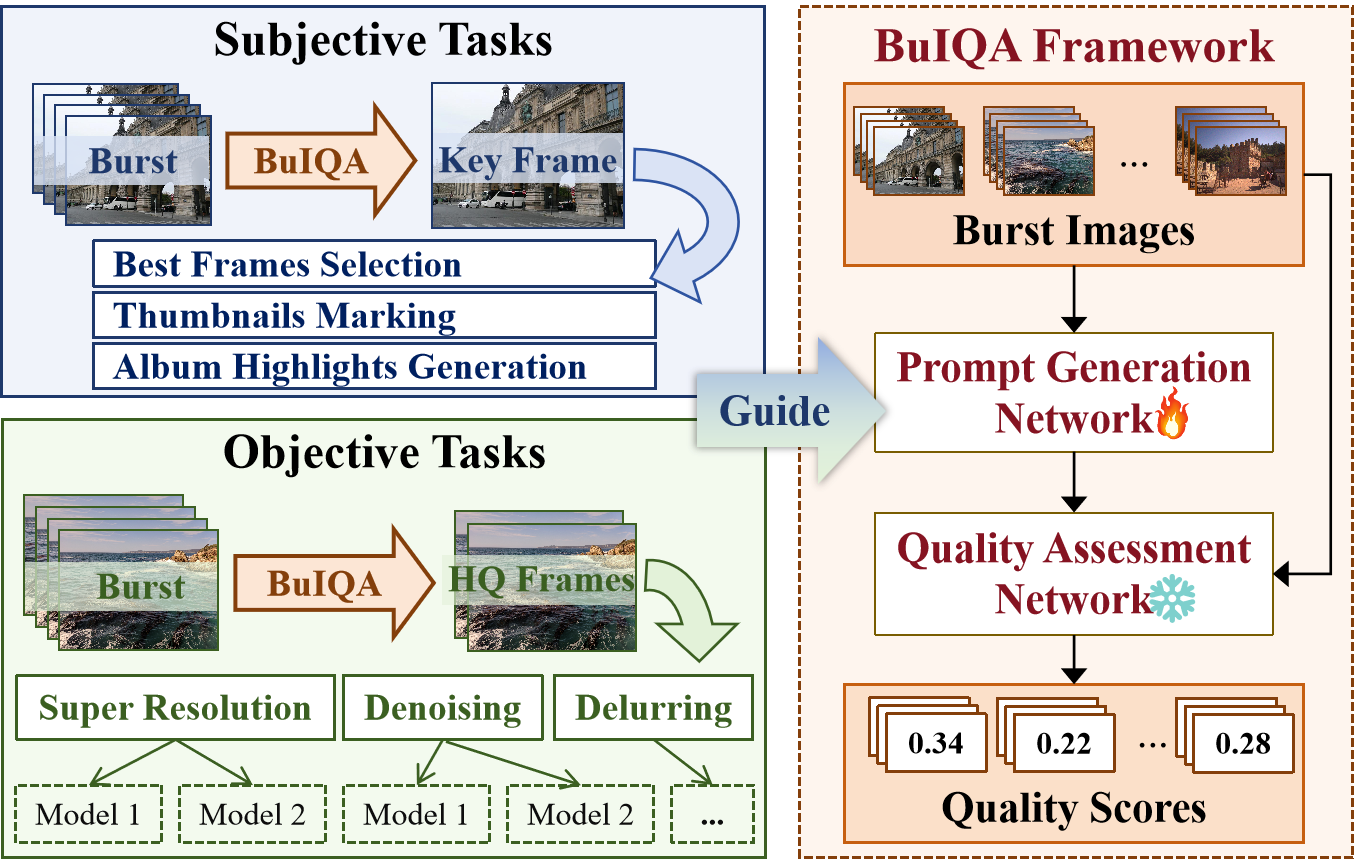} 
\caption{The illustration of the proposed BuIQA task and our unified framework based on task-aware prompt tuning. }
\label{fig:intro}
\end{figure}

Recently, Single Image Quality Assessment (SIQA) has received significant research interest. 
However, applying SIQA directly to burst images is challenging. First, compared with single image, the difference between burst frames is subtle and even visually imperceptible. Traditional SIQA approaches tailored for independent images are unable to assign the discriminative quality scores for consecutive frames. Second, as shown in Fig.\ref{fig:intro}, burst images serve a wide range of downstream tasks, in which the quality standards vary a lot. For example, aesthetic factors dominate the task of highlight generation, while fidelity and detail preservation are important to denoising. As a result, existing SIQA approaches are constrained by standardized evaluation pipelines, limiting their flexibility and efficacy in addressing the diverse downstream tasks of burst images. In this study, we propose a unified BuIQA framework with a task-aware prompt-tuning strategy, enabling fast adaptation to specific downstream tasks for more reliable quality assessment. Note that Video Quality Assessment (VQA) is also tailored for multiple frames as BuIQA, but it outputs overall video quality without discriminative score for each frame.

In this paper, we establish the first BuIQA dataset for multiple objective and subjective downstream tasks, comprising $7,346$ burst sequences. Inspired by the findings on the dataset, we propose a unified BuIQA framework for multiple downstream tasks, in a task-aware prompt-tuning manner. As shown in Fig.\ref{fig:intro}, the proposed framework consists of a tunable prompt generation network and a quality assessment network. 
The prompt generation network includes a Task-Driven Prompt Generation (TPG) module, which learns task-specific priors via heterogeneous knowledge distillation. The quality assessment network then leverages these prompts to guide task-aware feature extraction and multi-scale attention, enabling accurate quality evaluation of burst frames. The main contributions are three-fold:

\begin{itemize}
\item We propose a novel task of BuIQA with a new benchmark of $7,346$ burst sequences, which contains a total of $45,827$ images and $191,572$ quality score annotations for multiple downstream tasks. 
\item We propose a new task-aware prompt-tuning approach for BuIQA, showcasing the excellent quality assessment performance and generalization ability over objective and subjective downstream tasks. 
\item We introduce a prompt generation network with heterogeneous knowledge distillation for learning the task priors, and a task-aware quality assessment network for enabling task-driven BuIQA. 
\end{itemize}

\section{Related Work}
\noindent{\textbf{Burst Image Processing}}
Burst image processing refers to the technique of merging multiple frames from a burst sequence to generate a single high-quality output, which has been widely adopted in various computer vision tasks. For instance, several studies \cite{ehret2019joint,liu2023joint,guo2025compressed} utilize burst images with varying exposures to achieve high dynamic range imaging. Similar approaches \cite{bpa,monod2021analysis} leverage inter-frame context to suppress the noise in the current frame. 
Recently,  an attention-based fusion \cite{burstSR} and Swin Transformer  \cite{bsrt} architectures are respectively introduced for super-resolving burst images. However, the existing approaches process the whole burst sequence without considering the varying quality of individual frames, which limits efficiency and substantially degrades performance.

\noindent{\textbf{Visual Quality Assessment}}
Visual quality assessment is typically categorized into three main approaches: Full Reference (FR) \cite{kim2017friqa,wu2023friqa}, Reduced Reference (RR) \cite{liu2017rriqa,min2018rriqa}, and No Reference (NR) \cite{lin2018nriqa,liu2017nriqa,vqa3,VQA4}.
For instance, FRIQA \cite{wu2023friqa} is proposed for FR IQA by integrating low-level and high-level feature fusion. Similarly, a RR IQA model is presented based on the free-energy principle to improve image quality evaluation  \cite{liu2017rriqa}. 
Besides, Hallucinated-IQA \cite{lin2018nriqa}, a NR IQA method, leverages adversarial learning  for quality evaluation. However, the above quality assessment approaches cannot be directly applied to BuIQA.

\noindent{\textbf{Prompt-tuning}}
Prompt-tuning is a transfer learning technique that adapts pre-trained models to new tasks by optimizing a small set of parameters, mainly into two categories: explicit \cite{radford2019language,2021Learning,wang2023images} and implicit prompts \cite{jiang2024prompt,zhou2022coop,zhou2022cocoop,jia2022vpt}.
Explicit prompts rely on observable task-specific inputs, such as text or images. For example, GPT-2 \cite{radford2019language} performs various natural language processing tasks via textual prompts without task-specific training. Wang \textit{et al.} apply CLIP to assess image aesthetics using natural language supervision \cite{clipiqa}. In contrast to explicit prompts, VPT \cite{jia2022vpt}  introduces learnable prompt tokens to adapt vision transformers to specific tasks with minimal overhead. For BuIQA, implicit prompts are more reasonable due to the difficulty of defining the explicit evaluation criteria for diverse downstream tasks.

\section{Dataset and Analysis}
In this paper, a BuIQA dataset is established for multiple downstream tasks, consisting of two sub-datasets: Burst Image Objective Quality Assessment (BI-OQA) and Burst Image Subjective Quality Assessment (BI-SQA). In total, our BuIQA dataset comprises $7,346$ burst sequences with $45,827$ images and $191,572$ annotated quality scores. 

\subsection{Dataset Establishment}

\begin{figure*}[]
\centering
\includegraphics[width=1\linewidth]{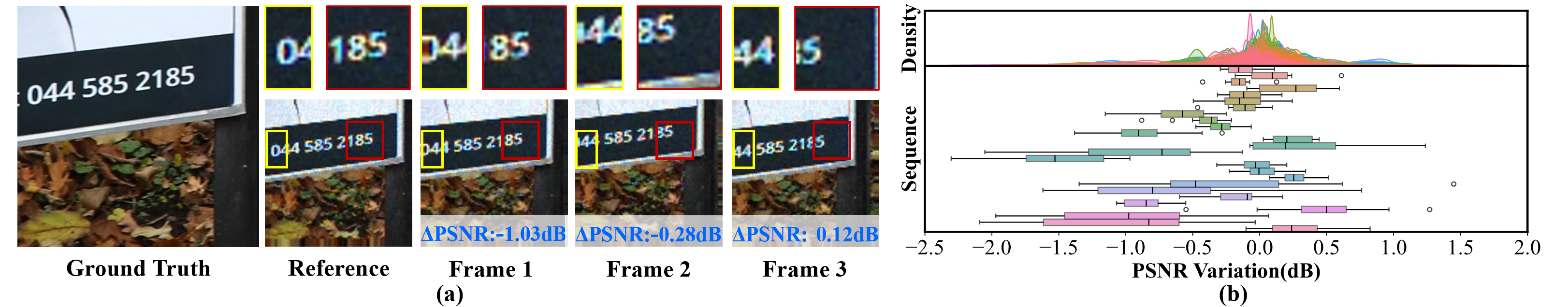} 
\caption{(a)  A burst sequence from our BI-OQA dataset. Blue values indicate PSNR variations when the corresponding frame is excluded from the input  sequence. (b) The boxplots and density curves of the PSNR variations when removing a frame from the sequence. The boxes and curves in different colors present the results of different sequences, and the black dots are outliers. }
\label{fig:ann2}
\end{figure*}

\textbf{BI-OQA}: 
In BI-OQA, we first collect real-world burst image datasets, BurstSR \cite{burstSR} and HDR+ \cite{hdr+} for two denoising and super-resolution, respectively.
After that, we apply downstream task models to infer the relative importance of different frames within a burst sequence, which are used as ground-truth quality scores. Specifically, 4 benchmark models for denoising, \ie, HDR21 \cite{monod2021analysis}, INN \cite{bpa}, DBD \cite{deep-burst-denoise} and BPN \cite{BPN}, and 4 for super-resolution,  \ie, EBSR \cite{ebsr}, BSRT \cite{bsrt}, DBSR \cite{burstSR} and BIP \cite{BIP}, are adopted as different downstream scenarios, since each benchmark model has its specific evaluation criterion. Moreover, to ensure consistency across tasks, we follow DBSR \cite{burstSR} and further synthesize 1,204 RAW burst sequences (14 frames each) for all the downstream scenarios.
Subsequently, frame-level scores are obtained via comparative experiments.
Finally, our BI-OQA contains a total of 2,237 burst sequences with 30,543 images and 176,288 annotations.

\noindent{\textbf{BI-SQA}}: 
The BI-SQA dataset is constructed by collecting and refining burst sequences and subjective quality scores from two datasets, \ie, Photo Triage \cite{auto-traige} and SPAQ \cite{fang2020perceptual}. 
Specifically, 4,175 burst sequences with 11,314 burst images, as well as corresponding quality scores are extracted from the training and validation sets of the Photo Triage dataset.
SPAQ, which contains individually annotated smartphone images, is grouped visually similar samples to form 934 burst sequences with 3,970 images. The original quality scores are normalized to the range [0, 1] for consistency.
Finally, the BI-SQA dataset contains a total of 5,109 burst sequences with 15,284 images.

\subsection{Data Analysis}
\textit{Findings 1: The importance of different frames in a burst sequence varies significantly for the downstream tasks.}

\textit{Analysis:} We evaluate frame impact by computing PSNR variation when removing each input.
Fig.~\ref{fig:ann2}(a) presents an example by the EBSR model \cite{ebsr}. 
We can observe that frame 3 fails to capture the digit ``0" due to positional displacement, and shows a certain level of degradation for the digit ``5".
Consequently, excluding frame 3 improves performance, while removing frames 1–2 degrades it.
Then, we conduct an analysis on the entire BI-OQA dataset.
The results are shown in Fig.~\ref{fig:ann2}(b), where the upper and lower panels depict the density distribution and the boxplot of PSNR variation. Quantitatively, 40\% of the frames are redundant ($\le$ 0.1 dB change),  25\% cause severe drops (up to 4 dB) and should be retained, while discarding 35\% low-quality frames yields slight gains. 

\textit{Findings 2: The ground-truth quality scores of burst frames remain consistent across sequences of different lengths.}

\begin{figure}[] 
\centering
{\includegraphics[width=1\linewidth]{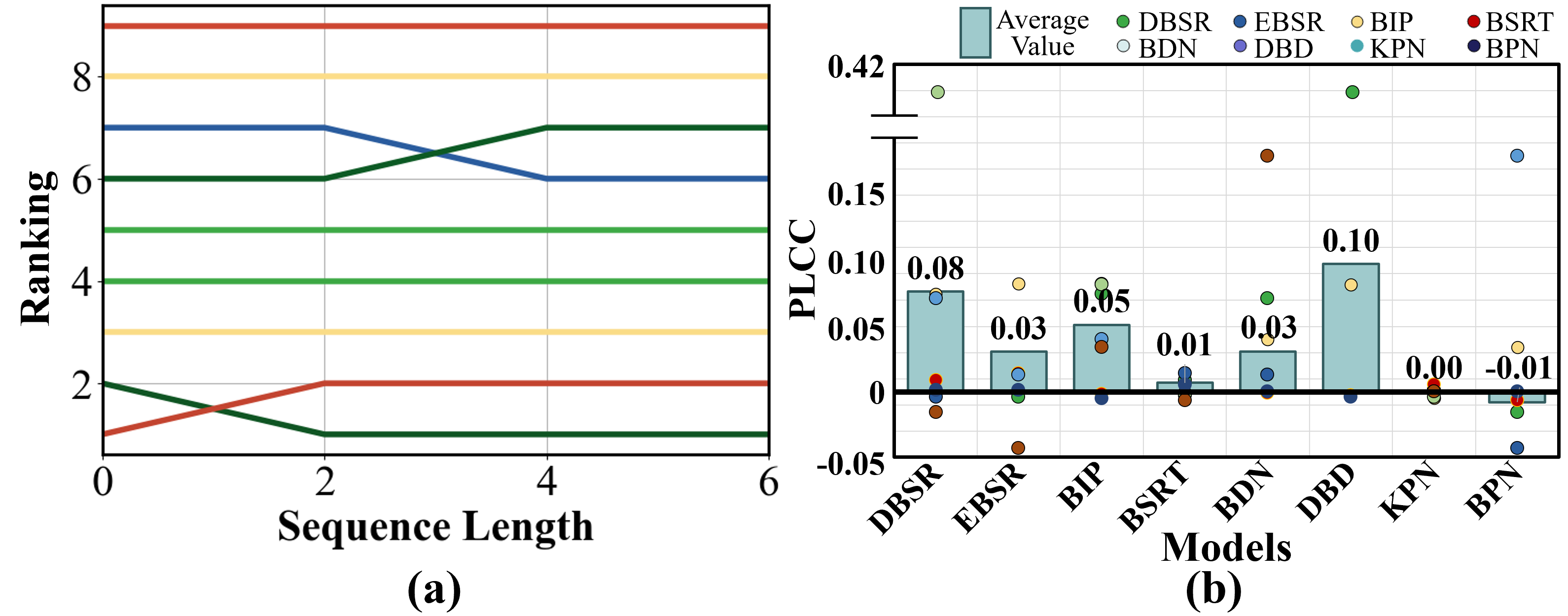}
	} 
\caption{
(a) Swimlane diagrams of sequence length and quality ranking, in which different colors represent the specific burst frames. (b) Histogram of PLCC among quality scores generated by different models and downstream tasks. }
\label{fig:ana_3}
\end{figure}

\textit{Analysis:} We investigate whether frame quality scores remain stable under varying sequence lengths. 
The quality rank of each frame is compared for visualization and analysis.
As shown in Fig. \ref{fig:ana_3}(a), most frame rankings are unchanged despite changes in input length, with only a few exhibiting minor shifts.
Furthermore, we calculate the PLCC values between ranking pairs under different sequence lengths. 
The PLCC results are over 0.7 for all pairs of our BI-OQA dataset, indicating a strong consistency.
The results indicate that the ground-truth quality scores of the burst frames are robust to changes in sequence length, reinforcing the reliability of quality score annotations in our dataset.

\textit{Findings 3: The objective quality score of burst frame varies significantly across different downstream tasks and models.}

\begin{figure*}[] 
	\centering
{\includegraphics[width=0.9\linewidth]{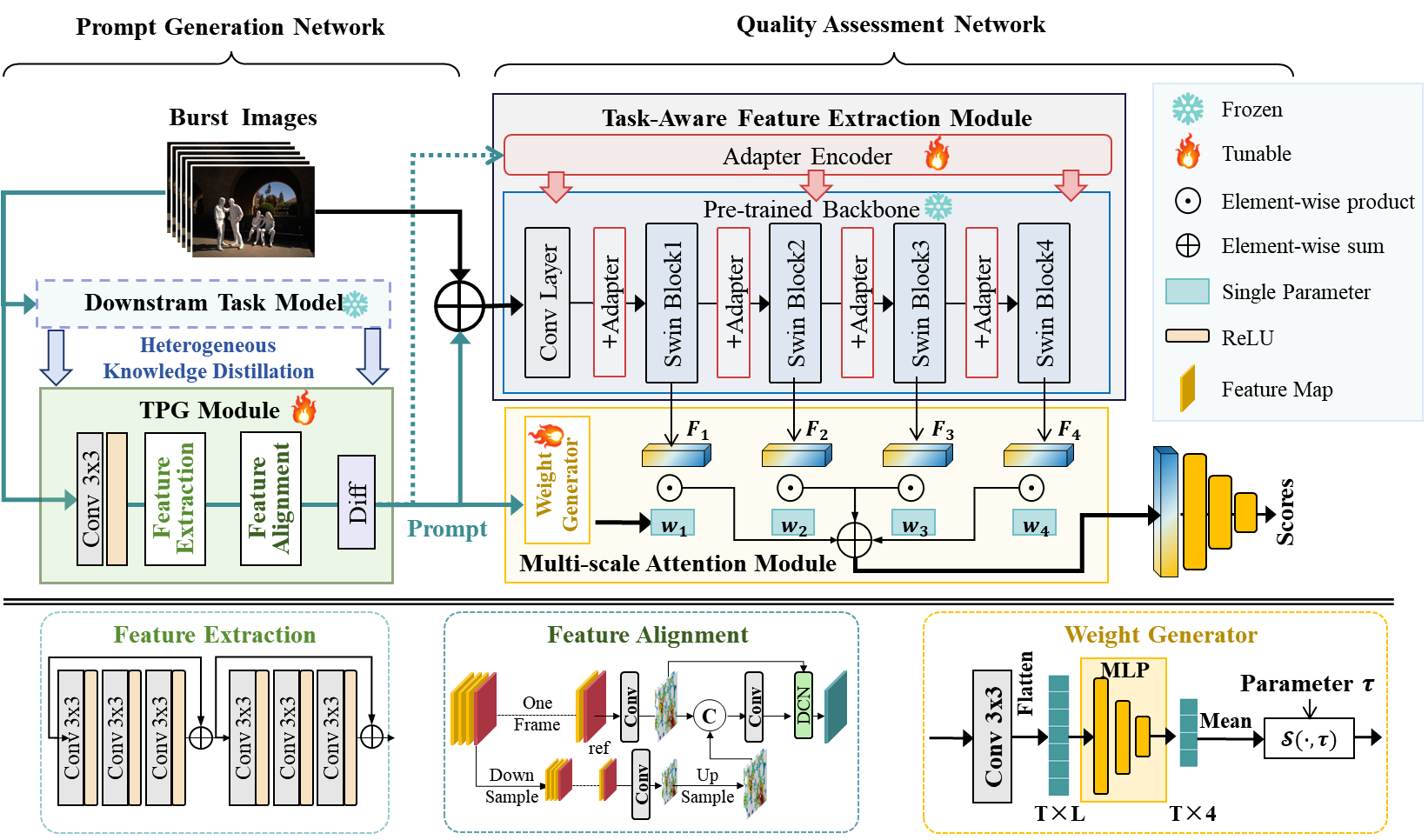}
	} 
	\caption{Illustration of our proposed framework.
		}
        \label{approach}
\end{figure*}

\textit{Analysis:} Based on the BI-OQA dataset, we explore the consistency of objective quality scores across different downstream tasks and model architectures. 
We first obtain the quality scores for each burst sequence through different models: KPN \cite{KPN}, BDN \cite{burstorm}, DBD \cite{deep-burst-denoise} and BPN \cite{BPN} for denoising task; and BIP \cite{BIP}, EBSR \cite{ebsr}, BSRT \cite{bsrt}, DBSR \cite{burstSR} for super-resolution task. 
Then, we compute the PLCC values between the quality scores from each pair of models. 
The results are shown in Fig. \ref{fig:ana_3}(b), from which we can observe that the average PLCC values of all pairs are less than 0.1.
The above results demonstrate that the objective quality score of a burst frame is highly correlated with downstream tasks and models.

\section{The Proposed Approach}
In this section, we propose a unified framework for BuIQA.
As illustrated in Fig. \ref{approach}, our framework consists of a prompt generation network and a quality assessment network. 
For prompt generation, a TPG module is constructed to generate task-driven prompt $\textbf{P}$ based on the input burst sequence $\textbf{B}$. Specifically, for objective downstream tasks, the TPG module learns priors of the tasks via knowledge distillation.
Note that for subjective tasks, it learns the subtle difference between frames of burst sequence, without the need of distillation. 
Given the generated prompt $\textbf{P}$ and the burst sequence  $\textbf{B}$, the quality assessment network first extracts task-specific burst features through the proposed task-aware feature extraction module, which is composed of a frozen pretrained backbone and a learnable adapter encoder. 
In this way, we can effectively adapt the knowledge-rich backbone to extract features well aligned with the downstream tasks.
Subsequently, an multi-scale attention module is devised to transform the prior information in the prompt into multi-scale attention weights, which are applied to the multi-scale features to produce refined feature representation $\mathrm{\mathbf{F_{fused}}}$.
Finally, a multi-layer perceptron (MLP) is employed on $\mathrm{\mathbf{F_{fused}}}$ to predict the task-aware quality scores $\mathrm{\hat{\mathbf{S}}}$ for all frames of the input burst sequence.

\subsection{Prompt Generation Network}
We propose the TPG Module to learn prior information of downstream tasks. 
Given an input burst sequence $\textbf{B} \in \mathbb{R} ^{T\times H\times W\times 3}$, the TPG Module generates a task-driven prompt $\textbf{P}\in \mathbb{R} ^{T\times H\times W\times 3}$, enabling the following quality assessment network to quickly adapt to specific downstream tasks. Specifically, the TPG module first extracts shallow features  $\textbf{Fea}_1$ from the input burst sequence using a convolutional layer followed by a ReLU activation.
Then, these features are passed through a deep feature extraction module, which is composed of two residual blocks, to capture high-level features $\textbf{Fea}_2$:
\begin{equation} \label{Eq:Fea2}
\mathbf{Fea}_2 = E_{FE}(\mathbf{Fea}_1).
\end{equation}
To further emphasize the subtle differences among burst images, we devise a feature alignment module to align each frame with the feature of the reference frame:
\begin{equation} \label{Eq:Fea3}
\textbf{Fea}_3 = E_{FA}
(Concat(\textbf{Fea}_2,\textbf{Fea}_2^{ref})))). 
\end{equation}
After that, we highlight the differences between each frame and the
reference frame through a differential operation, where the aligned feature $\textbf{Fea}_3$ are subtracted from the reference  feature $\textbf{Fea}_3^{ref}$:
\begin{equation} \label{Eq:Prompt}
\textbf{P} = \textbf{Fea}_3-\textbf{Fea}_3^{ref}. 
\end{equation}
Here, $\textbf{P}$ is taken as the output prompt of the TPG module, which embeds the subtle difference among frames of burst sequence. The prompt $\textbf{P}$ is then used to guide our quality assessment network for predicting quality scores of various downstream tasks.

\begin{figure}[] 
	\centering
{\includegraphics[width=1\linewidth]{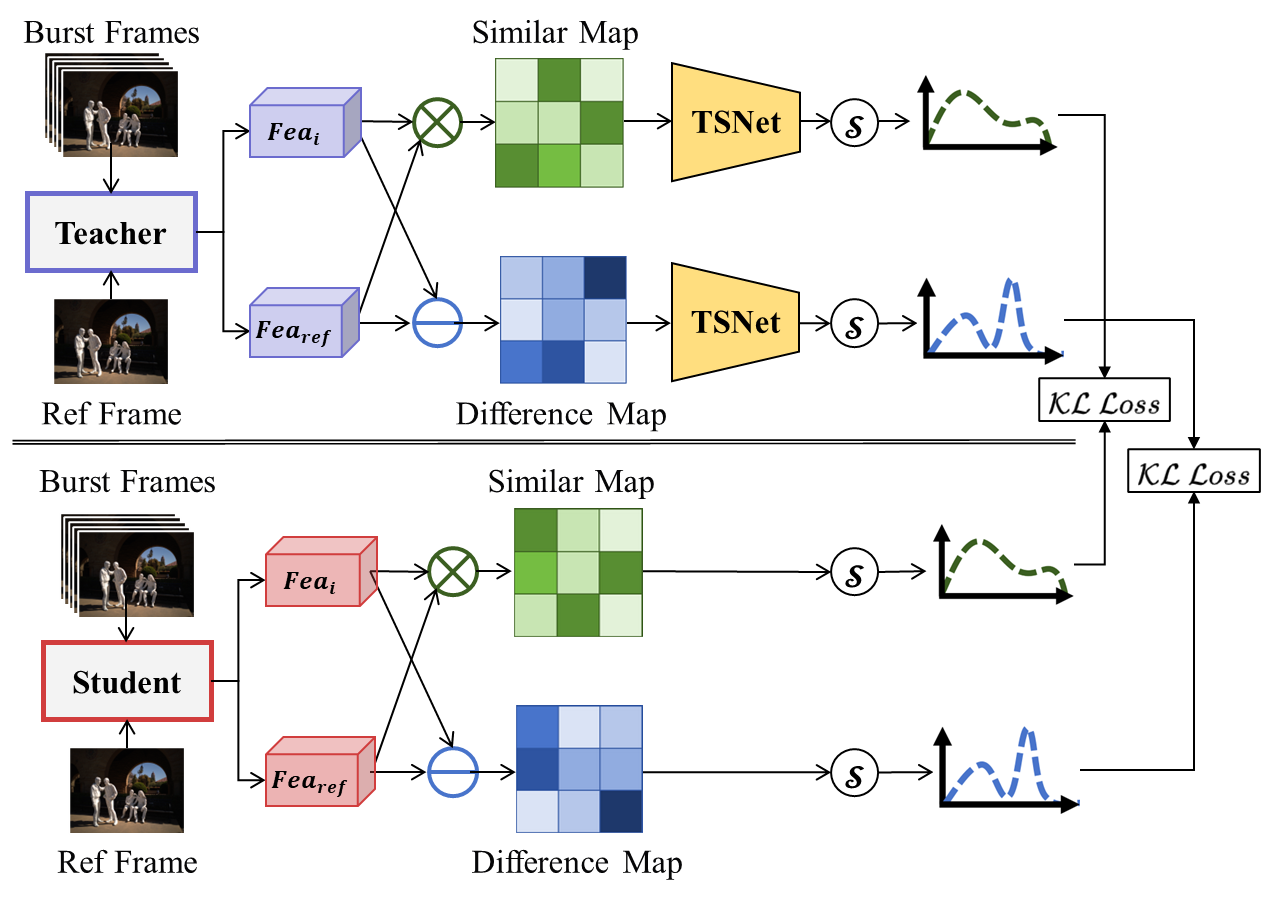}
	} 
	\caption{The illustration of our knowledge distillation.
		} 
        \label{dist}
\end{figure}

\noindent\textbf{Heterogeneous Knowledge Distillation.}
\textit{Finding 3} indicates that the quality scores of different downstream models show weak correlation, even within the same task. Therefore, we propose to leverage knowledge distillation to learn priors from downstream task models. 
In traditional homogeneous knowledge distillation, the student model usually has similar structure with the teacher model. However, our framework requires the unified student model to learn from diverse downstream task teacher models with different architectures. 
To solve this, we propose a relational heterogeneous knowledge distillation approach, as illustrated in Fig. \ref{dist}. 
Specifically, we employ a similar map $\textbf{Map}_{\text{sim}}$ to represent the shared components, and a difference map $\textbf{Map}_{\text{diff}}$ to represent complementary differences between the features from teacher and student models:
\begin{align} \label{Eq:Dist}
\textbf{Map}_{\text{sim}} &= \textbf{Fea}_{\text{ref}} \odot \textbf{Fea}_{\text{i}}, \\
\textbf{Map}_{\text{diff}} &= \textbf{Fea}_{\text{ref}} - \textbf{Fea}_{\text{i}}.
\end{align}
Here, $\textbf{Fea}_i$ denotes the feature of $i$-th frame. 
Since the architecture of the teacher model is variable, inspired by Hint. \textit{et al.} \cite{romero2014fitnets}, we map the teacher features to the student feature space using TSNet:
\begin{equation} \label{Eq:map}
\mathbf{{Map_x^t}} = TSNet(\mathbf{\hat{Map_x^t}}), \mathbf{x} \in \{\text{sim}, \text{diff}\},
\end{equation}
where $\mathbf{t}$ indicates the teacher model. Then, all maps are normalized via a softmax function to obtain probability distributions:
\begin{equation} \label{Eq:diff}
\mathbf{{D_x^y}} =Softmax(\mathbf{{Map_x^y}}), \mathbf{y}\in\{\mathbf{t},\mathbf{s}\}. 
\end{equation}
Here, $\mathbf{s}$ indicates the student model. 
Finally, we adopt the Kullback-Leibler (KL) divergence loss function to supervise the knowledge distillation process:
\begin{equation} \label{Eq:kl}
\mathcal{L}_{\text{Dist}} =\sum_{\mathbf{x\in\{sim,diff\}}}\ \mathbf{{D_x^t}}\cdot \log(\frac{\mathbf{{D_x^t}}}{\mathbf{{D_x^s}}})+(1-\mathbf{{D_x^t}})\cdot \log(\frac{1-\mathbf{{D_x^t}}}{1-\mathbf{{D_x^s}}}).
\end{equation}
Note that the computation in Eq. (\ref{Eq:kl}) is conducted in an element-wise manner.

\subsection{Quality Assessment Network}

The Quality Assessment Network consists of two key components: a task-aware feature extraction module and a multi-scale attention module. 
The former consists of a pre-trained backbone and an adapter encoder for task-aware feature extraction, and the latter one is devised to generate multi-scale attention weights for feature refinement.
More details of the two modules are discussed as follows.

\noindent\textbf{Task-aware feature extraction module.} 
We adopt Swin Transformer as the pre-trained backbone.
To enhance its task-aware adaptability, we introduce an adapter encoder to transfer the backbone for specific tasks. 
Specifically, the adapter encoder learns task-specific representations from the prior information in the prompt $\mathbf{P}$, and its outputs are combined with the multi-scale features of the backbone and fed into the subsequent layer:
\begin{equation} \label{Eq:Fi}
\mathbf{F}_i = Block_{i}(\mathbf{F}_{i-1}+Adapter_i(\mathbf{P})),
\end{equation}
where \( i \) represents  \( i-{th} \) block in the pre-trained backbone.
This process yields four-level of features \( \mathbf{F}=\{\mathbf{F}_1,\mathbf{F}_2,\mathbf{F}_3,\mathbf{F}_4\} \) at different scales, capturing both low-level details (e.g., noise and blur) and high-level semantics (e.g., composition and tone).

\begin{table*}[h]
\centering
\footnotesize
{
\begin{tabular}{c|c|c|c|c|c|c|c|c|c}
    \toprule
    \hline
&
  &
  \multicolumn{4}{c|}{{\textbf{Denoising}}} &
  \multicolumn{4}{c}{{\textbf{Super-resolution}}} \\  
\multirow{-2}{*}{{\textbf{Metrics}}}&\multirow{-2}{*}{{\textbf{Approach}}}&
  \multicolumn{1}{c|}{{\textbf{~HDR21~}}} &
  \multicolumn{1}{c|}{{\textbf{~~INN~~}}} &
  \multicolumn{1}{c|}{{\textbf{~~DBD~~}}} &
  \multicolumn{1}{c|}{{\textbf{~~BPN~~}}} &
  \multicolumn{1}{c|}{{\textbf{~~EBSR~~}}} &
  \multicolumn{1}{c|}{{\textbf{~~BSRT~~}}} &
  \multicolumn{1}{c|}{{\textbf{~~DBSR~~}}} &
  \multicolumn{1}{c}{{\textbf{~~BIP~~}}} \\
 \hline
&
  {Baseline} &
   0.266 & 
  {0.139} &
   0.352 &
   0.104&
  {0.087} &
   \underline{0.077} &
   \underline{0.051} &
   \underline{0.074} \\
&
  {PAU} &
   0.397 & 
  {0.139} &
   \underline{0.601} &
   \underline{0.220} &
   \underline{0.143} &
   0.028 &
   0.041 &
   0.010 \\
&
  {SPAQ} &
  {0.276} &
  {0.177} &
  {0.175} &
  {0.108} &
  {0.087} &
  {0.035} &
  {0.035} &
  {0.020} \\
&
  {PAUQA} &
  {0.369} &
  {0.196} &
  {0.317} &
  {0.038} &
  {0.111} &
  {0.023} &
  {0.034} &
  {0.061} \\
&
  {ELTA} &
  {0.197} &
  {0.041} &
  {0.079} &
  {0.149} &
  {0.017} &
  {0.015} &
  {0.021} &
  {0.030} \\
  &
  {ESFD} &
  {0.206} &
   \underline{0.406} &
  {0.098} &
  {0.137} &
   0.002 &
  {0.004} &
  {0.003} &
  {0.005} \\
&
  {FasterVQA} &
  {0.247} &
  {0.141} &
  {0.060} &
  {0.218} &
  {0.061} &
  {0.043} &
  {0.039} &
  {0.033} \\
&
  {KVQ} & \underline{0.421}&0.098&0.352&0.026&0.053&0.007&0.034&0.018\\
\multirow{-9}{*}{{\textbf{R0}}} &
  {\textbf{Ours}} &
  {\textbf{0.810}} &
  {\textbf{0.447}} &
  {\textbf{0.768 }}&
  {\textbf{0.367}} &
  {\textbf{0.486}} &
  {\textbf{0.269}} &
  {\textbf{0.282}} &
  {\textbf{0.292}} \\ \hline
&
  {Baseline} &
  {0.230} &
  {0.046} &
  {0.404} &0.107&
  {0.240} &
  {0.124} &
  {0.150} &
  {0.236} \\
&
  {PAU} &
   0.231 &
  {0.074} &
   \underline{0.602} &
  {0.257} &
   \underline{0.313} &
   \underline{0.157} &
   0.154&
  {0.197} \\
&
  {SPAQ} &
   0.508 &
  {0.150} &
  {0.233} &
  {0.186} &
  {0.274} &
   0.098 &
  {0.132} &
  {0.228} \\
&
  {PAUQA} &
  {0.391} &
  {0.244} &
  {0.333} &
  {0.202} &
  {0.259} &
  {0.096} &
  {0.137} &
  {0.163} \\
&
  {ELTA} &
  {0.260} &
  {0.136} &
  {0.091} &
   0.277 &
  {0.201} &
  {0.110} &
  {0.107} &
   \underline{0.245} \\
   &
  {ESFD} &
  {0.284} &
   \underline{0.438} &
  {0.100} &
  {0.234} &
   0.211 &
  {0.083} &
  {0.108} &
  {0.034} \\
&
  {FasterVQA} &
  {0.261} &
  {0.152} &
  {0.063} &
   \underline{0.338} &
  {0.239} &
  {0.124} &
  {0.147} &
   0.170 \\
&
  {KVQ} &\underline{0.565}&0.100&0.562&0.099&0.227&0.105&\underline{0.170}&0.202
  \\
\multirow{-9}{*}{{\textbf{R0.02}}} &
  {\textbf{Ours}} &
  {\textbf{0.862}} &
  {\textbf{0.487}} &
  {\textbf{0.756}} &
  {\textbf{0.427}} &
  {\textbf{0.553}} &
  {\textbf{0.315}} &
  {\textbf{0.363}} &
  {\textbf{0.350}} \\ \hline
&
  {Baseline} &
  {0.283} &
  {-0.003} &
  {0.483} &
  0.137&
  {0.389} &
  {0.154} &
  {0.262} &
   \underline{0.298} \\
&
  {PAU} &
   0.532 &
  {0.090} &
   \underline{0.611} &
  {0.243} &
   \underline{0.445} &
   \underline{0.201} &
   0.270 &
  {0.288} \\
&
  {SPAQ} &
   \underline{0.691} &
  {0.167} &
  {0.288} &
  {0.188} &
  {0.423} &
   0.151 &
  {0.242} &
  {0.271} \\
&
  {PAUQA} &
  {0.372} &
  {0.264} &
  {0.356} &
  {0.235} &
  {0.426} &
  {0.148} &
  {0.271} &
  {0.213} \\
&
  {ELTA} &
  {0.321} &
  {0.117} &
  {0.097} &
  {0.331} &
  {0.408} &
  {0.138} &
  {0.211} &
  {0.253} \\
  &
  {ESFD} &
  {0.310} &
   \underline{0.441} &
  {0.082} &
  {0.290} &
   0.366 &
  {0.128} &
  {0.225} &
  {0.230} \\
&
  {FasterVQA} &
  {0.252} &
  {0.141} &
   0.090 &
   \underline{0.353} &
  {0.407} &
  {0.169} &
   \underline{0.277} &
   0.270 \\
&
  {KVQ} & 0.645&0.161&0.606&0.250&0.375&0.159&0.271&0.123
  \\
\multirow{-9}{*}{{\textbf{R0.05}}} &
  {\textbf{Ours}} &
  {\textbf{0.934}} &
  {\textbf{0.497}} &
  {\textbf{0.743}} &
  {\textbf{0.435}} &
  {\textbf{0.658}} &
  {\textbf{0.337}} &
  {\textbf{0.390}} &
  {\textbf{0.489}} \\ \hline
&
  {Baseline} &
  {0.318} &
  {0.036} &
  {0.515} &
  0.165&
  {0.510} &
  {0.219} &
  {0.278} &
   \underline{0.275} \\
&
  {PAU} &
   0.656 &
  {0.130} &
   0.542 &
  {0.321} &
  {0.528} &
   \underline{0.253} &
   0.327 &
  {0.244} \\
&
  {SPAQ} &
   \underline{0.795} &
  {0.178} &
  {0.230} &
  {0.161} &
   \underline{0.561} &
   0.210 &
  {0.302} &
  {0.241} \\
&
  {PAUQA} &
  {0.436} &
  {0.256} &
   0.366 &
  {0.337} &
  {0.529} &
  {0.186} &
   \underline{0.334} &
  {0.203} \\
&
  {ELTA} &
  {0.337} &
  {0.083} &
  {0.148} &
  {0.334} &
  {0.525} &
  {0.214} &
  {0.268} &
  {0.226} \\
  &
  {ESFD} &
  {0.417} &
   \underline{0.473} &
  {0.090} &
  {0.295} &
   0.525 &
  {0.154} &
  {0.285} &
  {0.005} \\
&
  {FasterVQA} &
  {0.303} &
  {0.141} &
  {0.111} &
   \underline{0.436} &
  {0.529} &
  {0.232} &
  {0.324} &
   0.222 \\
&
  {KVQ} & 0.714&0.197&\underline{0.578}&0.376&0.514&0.199&0.329&0.177
  \\
\multirow{-9}{*}{{\textbf{R0.1}}} &
  {\textbf{Ours}} &
  {\textbf{0.969}} &
  {\textbf{0.523}} &
  {\textbf{0.661}} &
  {\textbf{0.483}} &
  {\textbf{0.718}} &
  {\textbf{0.373}} &
  {\textbf{0.488}} &
  {\textbf{0.565}}\\ \hline \bottomrule
\end{tabular}}
\caption{The performance of our and compared approaches. The best and second best results are in \textbf{bold} and \underline{underlined}.}
  \label{tab:result}
\end{table*}

\noindent\textbf{Multi-scale attention module.}
The required features for BuIQA vary at different tasks: assessments for subjective tasks focus on both low-level (e.g., noise, blur) and high-level attributes (e.g., composition, color), while assessment for objective tasks rely more on pixel-level information due to minimal high-level variations across frames.
In this paper, we infer these differences based on the task-driven prompt.
Specifically, we design a multi-scale attention module, which takes the prompt as input and generates attention weights to dynamically fuses the multi-scale features. 
Specifically, the module generates 4 attention weight values, which are normalized via softmax to produce attention weights:
\begin{equation} \label{Eq:w}
\mathbf{w} = Softmax(E_{wg}(\mathbf{P})).
\end{equation}
The weighted features are then fused to generate refined representation:
\begin{equation} \label{Eq:fused}
\mathbf{F_{fused}} = \sum_{i=0}^4 \mathbf{w}_i\cdot \mathbf{F}_i.
\end{equation}
Finally, an MLP is applied on the refined representation to predict the final quality score $\mathbf{\hat{S}}$ of the sequence:
\begin{equation} \label{Eq:S}
\mathbf{\hat{S}} = MLP(\mathbf{F_{fused}}),
\end{equation}
where $\mathbf{\hat{S}}=\{\hat{S}_1,\hat{S}_2,...,\hat{S}_n\}$, $n$ is the length of the sequence.

\noindent\noindent\textbf{Margin Loss function}.
To further emphasize differences between burst frames, we employ a margin loss \cite{wang2019real} function to train our quality assessment network.
Specifically, the margin loss function is defined as:
\begin{equation} \label{Eq:pair}
\mathcal{L}_{\text{pair}}(i,j) = ReLU((S_i-S_j)-(\hat{S}_i-\hat{S}_j)),
\end{equation}
where $i$ and $j$ denote the indices of images from different groups, respectively. $S_i$ represents the ground truth score, $\hat{S}_i$ are the predicted score.
For example, if frame $i$ is of higher quality than frame $j$ (e.g., ${S}_i-{S}_j>0$), the loss penalizes incorrect rankings (e.g., $\hat{S}_i-\hat{S}_j<0$) and underestimated differences (e.g., $0< \hat{S}_i-\hat{S}_j < {S}_i-{S}_j$). 
\textit{Finding 2} reveals that only a small number of frames undergo ranking changes when using different sequence lengths, and these changes are limited to a small range. 

To address these minor variations, we introduce a ``grouping rank" approach. To be specific, frames with similar scores are grouped together, and their rankings are treated as equivalent, while frames without ranking changes retain their original ranks. 
This grouping approach ensures that slight score fluctuations do not disturb the overall ranking structure, thereby enhancing the robustness and reliability of quality assessment. 
Let $\mathbf{G}$ represents groups, and the final loss is computed as the mean of $\mathcal{L}_{\text{pair}}$ over all image pairs:
\begin{equation} \label{Eq:mrg}
\mathcal{L}_{\text{Mrg}} = \sum_{(i,j)\in \mathbf{Pair}} \mathcal{L}_{\text{pair}}(i,j)/N_P,
\end{equation}
where $\mathbf{Pair} =\{(i,j)|i\in \mathbf{G}_k,j \notin \mathbf{G}_k,k\in \mathbb{Z}^+, k\le N_G\}$. 
$N_G$ and $N_P$ are the lengths of 
 $\mathbf{G}$ and $\mathbf{Pair}$, respectively.
Finally, the overall loss can be formulated as:
\begin{equation} \label{Eq:loss}
\mathcal{L}_{\text{fnl}} = \alpha\mathcal{L}_{\text{Dist}}+\beta\mathcal{L}_{\text{Mrg}},
\end{equation}
where $\alpha$ and $\beta$ are hyper-parameters for balancing each individual loss.

\section{Experimental Results}
In this section, we conduct experiments to evaluate our method for BuIQA over both subjective and objective downstream tasks.
Specifically, both the proposed BI-OQA and BI-SQA datasets are randomly split into training and test sets at a ratio of 4:1.
To balance different loss terms, the hyper-parameters ${\alpha, \beta}$ are set to ${1, 10}$, respectively.
For training our model, the Adam optimizer is adopted for parameter optimization, and the initial learning rate is set to $1 \times 10^{-3}$.
Since BuIQA is a newly proposed task without comparative approaches, we first apply VGGNet-16 \cite{vggnet} as the baseline model, supervised by our margin loss. Moreover, we compare our approach  with 7 state-of-the-art quality assessment approaches, including 5 IQA approaches, \ie, PAU \cite{huang2022series}, SPAQ \cite{fang2020perceptual}, PAUQA \cite{PAUQA}, ELTA \cite{ELTA} and ESFD \cite{dong2025exploring}, and 2 VQA approaches, \ie, FasterVQA \cite{FasterVQA} and KVQ \cite{qu2025kvq}.
Note that all the baseline and compared approaches are trained and evaluated under the same settings as our approach. 

\subsection{Evaluation on Objective Tasks}
\begin{figure}[] 
	\centering
{\includegraphics[width=1.\linewidth]{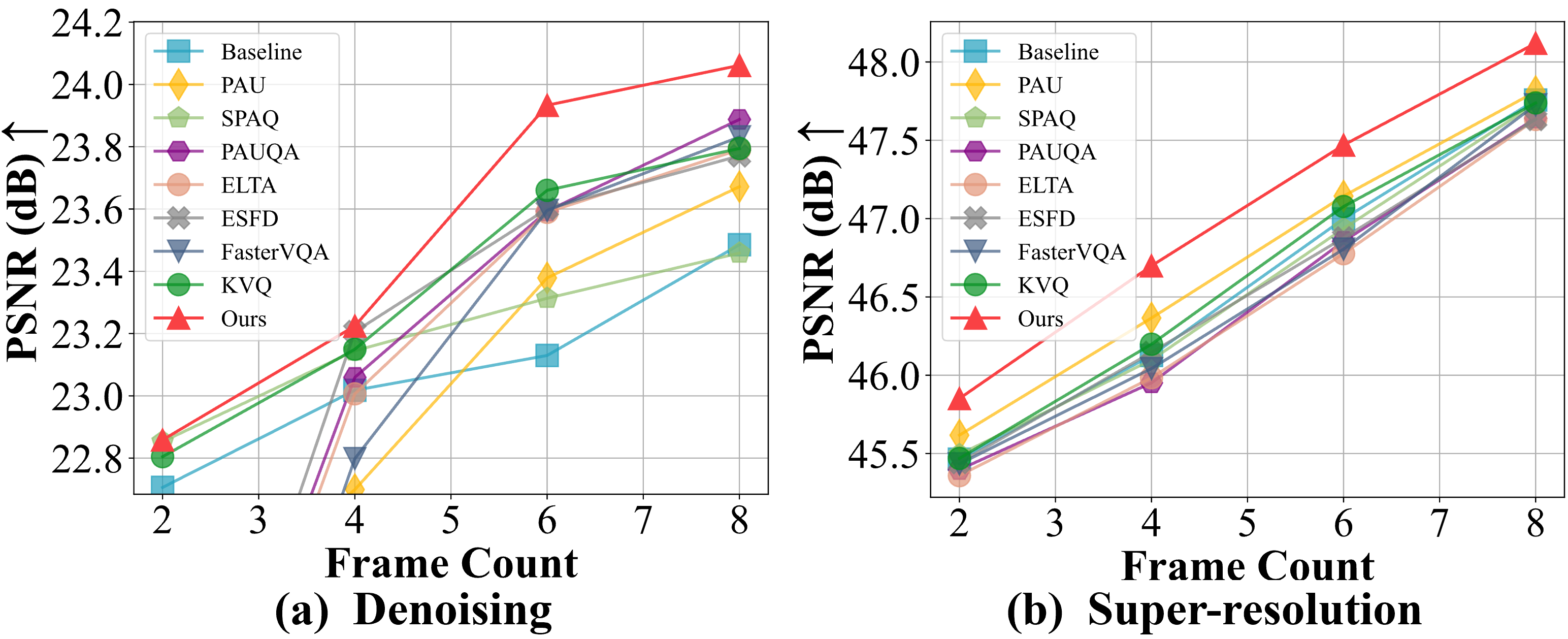}
	} 
	\caption{Performance comparison on downstream tasks.
		} 
        \label{fig:obj}
\end{figure}
We evaluate the quality assessment performance for objective tasks, \ie, denoising and super-resolution on the proposed BI-OQA dataset.
We leverage the SRCC to measure the  quality assessment performance.
Besides, recall that \textit{Finding 2} demonstrates that frames with comparable quality scores contribute almost equally to downstream tasks.
In other words, for the evaluation of BuIQA, it is reasonable to treat frames with comparable scores as the same importance level for downstream tasks.
Inspired by this, we employ a more flexible evaluation metric to introduce relaxed thresholds: R$0$, R$0.02$, R$0.05$, and R$0.1$. 
Taking R$0.02$ as an example, if the gap between the predicted and ground-truth quality scores is less than 0.02, it would be considered equivalent. 
The larger threshold indicates a more flexible evaluation metric. 
Tab.~\ref{tab:result} shows the results in BI-OQA, where our approach outperforms all compared approaches in terms of SRCC, under all 2 downstream tasks and 8 annotations by models. 
Specifically, for EBSR model of SR, our approach promotes SRCC by at least 0.343, 0.240, 0.213 and 0.157 under the relaxed thresholds of R$0$, R$0.02$, R$0.05$, and R$0.1$, respectively. Similar results can be found for denoising task.
The above results validate the effectiveness of our approach on BuIQA.

Besides, we also evaluate the practical utility of our BuIQA approach over the two downstream tasks, \ie, INN \cite{bpa} for denoising and EBSR \cite{ebsr} for super-resolution, via the performance gain in terms of PSNR.
Here, we select the burst frames according to the quality scores from our and compared approaches. 
Specifically, as can be seen in Fig.~\ref{fig:obj}, for frame count $M$, only the frames with top-$M$ predicted scores are input to the downstream models.
It shows that under different frame counts, our approach  outperforms almost all compared approaches in both the denoising and super-resolution tasks.
The above results validate the practical utility of our approach on downstream tasks.

\subsection{Evaluation on Subjective Tasks}

\begin{table}
\footnotesize
   \centering{
  \begin{tabular}{c|c|c|c}
    \toprule
    \hline
  \multicolumn{2}{c|}{\textbf{Approach}} & \textbf{Photo Traige} & \textbf{SPAQ}\\
    \midrule
    &  Baseline&  0.631$\pm$0.127&0.501$\pm$0.087 \\
    & PAU &  0.632$\pm$0.074& 0.524$\pm$0.079\\
    & SPAQ & 0.584$\pm$0.135 & 0.566$\pm$0.072\\
    & PAUQA & 0.639$\pm$0.124 &0.501$\pm$0.103\\
    & ELTA&0.680$\pm$0.127 &0.510$\pm$0.094 \\ 
    \multirow{-6}{*}{\textbf{IQA}}& ESFD & 0.665$\pm$0.128 &0.538$\pm$0.091 \\\hline
    & FasterVQA & 0.587$\pm$0.142 &0.503$\pm$0.093\\ 
    \multirow{-2}{*}{\textbf{VQA}}& KVQ &0.533$\pm$0.144 &0.472$\pm$0.112\\ 
    \hline
     \textbf{BuIQA}& \textbf{Ours} & \textbf{0.706$\pm$0.121}&\textbf{0.582$\pm$0.067} \\
    \hline
  \bottomrule
\end{tabular}}
\caption{Results of pairwise accuracy on BI-SQA dataset}
\label{tab:sub}
\end{table}

Here, we evaluate the quality assessment for subjective downstream tasks, \ie, subjective frame selection on BI-SQA dataset.
For evaluation, we leverage the widely used pairwise accuracy metric \cite{auto-traige}, which compares the scores of each frame pair for each burst image.
As can be seen in Tab.~\ref{tab:sub}, our approach outperforms all compared approaches on both the Photo Traige \cite{auto-traige} and SPAQ \cite{fang2020perceptual} datasets.
For instance, our approach achieves at least 0.026 and 0.016 accuracy promotion on the two datasets, respectively.
The above results validate the effectiveness of our approach on subjective downstream tasks in terms of IQA accuracy and frame selection.

\begin{table}[!t]
\centering
\footnotesize
{
\begin{tabular}{c|c|c|c|c}
\toprule
\hline
\multicolumn{4}{c|}{\textbf{Settings}}   &\textbf{Metrics} \\
\hline
\multicolumn{2}{c|}{\textbf{Prompt}} & \multicolumn{2}{c|}{\textbf{Quality assessment}}         & \multirow{2}{*}{\textbf{SRCC}} \\ \cline{1-4} 
TPG & $\mathcal{L}_{\mathrm{Dist}}$ & MS features & MS attention &                     \\
\hline
        &    &  \textbf{   $\checkmark$ }  & \textbf{   $\checkmark$ }                 & 0.500    
      \\
           \textbf{   $\checkmark$ } &      &   \textbf{   $\checkmark$ }    &    \textbf{   $\checkmark$ }                     & 0.643 \\
 \textbf{   $\checkmark$ }&   \textbf{   $\checkmark$ }&     \textbf{   $\checkmark$ }  &          &0.470   \\

   \textbf{   $\checkmark$ }  &\textbf{   $\checkmark$ }  &                       &           \textbf{   $\checkmark$ }                                    & 0.773  \\
   \hline
   \textbf{   $\checkmark$ }
                   &  \textbf{   $\checkmark$ } &      \textbf{   $\checkmark$ }    &                    \textbf{   $\checkmark$ }           & 0.818   \\
\hline
\bottomrule
    
\end{tabular}%
}
\caption{Ablation results of our approach.}
  \label{tab:ab}
\end{table}

\subsection{Ablation Studies}
We perform ablation experiments to assess the effectiveness of our prompt generation network and quality assessment network. For the prompt generation network, we test our model without two main components: (1)  TPG module, (2) distillation loss $\mathcal{L}_{\mathrm{Dist}}$. 
For the quality assessment network, we also test without two components: (1) multi-scale (MS) features in task-aware feature extraction module, (2) MS attention module.
In each case, the other network is kept fixed with its complete configuration. As can be seen in Tab.~\ref{tab:ab}, all settings result in performance drops. Notably, the disregard of distillation loss and MS attention degrades SRCC by at least 0.175 and 0.348, respectively. These results confirm the efficacy of each component in our design.

\section{Conclusion}
In this paper, we introduced the novel task of BuIQA, which evaluates the task-driven quality of individual frames within burst sequences, enabling effective frame selection for both subjective and objective downstream tasks. We established the first benchmark dataset for BuIQA. 
Based on comprehensive analysis on our dataset, we uncovered key insights into how frame quality affects downstream performance.
Guided by the findings, we proposed a unified BuIQA framework with a task-aware prompt-tuning strategy, integrating a prompt generation network and a quality assessment network for adaptive evaluation. Extensive experiments across downstream tasks demonstrated that BuIQA outperforms existing approaches and significantly boosts denoising and super-resolution performance.

\section*{Acknowledgements}
This work was supported by NSFC under Grants 62231002,62401027, 62372024, 62206011, 62450131, the Fundamental ResearchFunds for the Central Universities, and the Academic Excellence Foundation of BUAA for PhD Students.

\bibliography{aaai2026}

\end{document}